\documentclass[letterpaper, 10 pt, conference]{ieeeconf}  

\IEEEoverridecommandlockouts                              

\usepackage{subfigure}
\usepackage{pifont}
\usepackage{textcomp}
\usepackage{stfloats}
\usepackage{url}
\usepackage{verbatim}
\usepackage{graphicx}
\usepackage{cite}
\usepackage{multirow}
\usepackage{svg}
\usepackage[utf8]{inputenc}

\title{\LARGE \bf
DABI: Evaluation of Data Augmentation Methods Using Downsampling in Bilateral Control-Based Imitation Learning with Images
}

\author{ Masato Kobayashi$^\dag$$^{1,2*}$,Thanpimon Buamanee$^\dag$$^{2}$,  Yuki Uranishi$^{1,2}$
\thanks{$^{\dag}$Equal Contribution}
\thanks{$^{1}$Cybermedia Center, Osaka University, Osaka, 560-0043 Japan}
\thanks{$^{2}$Graduate School of Information Science and Technology, Osaka University, Osaka 560-0043, Japan, %
}}

\begin{document}

\maketitle
\thispagestyle{empty}
\pagestyle{empty}

\begin{abstract}
Autonomous robot manipulation is a complex and continuously evolving robotics field. This paper focuses on data augmentation methods in imitation learning. Imitation learning consists of three stages: data collection from experts, learning model, and execution. However, collecting expert data requires manual effort and is time-consuming. Additionally, as sensors have different data acquisition intervals, preprocessing such as downsampling to match the lowest frequency is necessary.
Downsampling enables data augmentation and also contributes to the stabilization of robot operations.
In light of this background, this paper proposes the Data Augmentation Method for Bilateral Control-Based Imitation Learning with Images, called "DABI". DABI collects robot joint angles, velocities, and torques at 1000 Hz, and uses images from gripper and environmental cameras captured at 100 Hz as the basis for data augmentation.
This enables a tenfold increase in data. 
In this paper, we collected just 5 expert demonstration datasets.
We trained the bilateral control Bi-ACT model with the unaltered dataset and two augmentation methods for comparative experiments and conducted real-world experiments.
The results confirmed a significant improvement in success rates, thereby proving the effectiveness of DABI.
For additional material, please check: \url{https://mertcookimg.github.io/dabi}
\end{abstract}

\section{INTRODUCTION}
The paradigm of robot control has undergone significant changes, with a particular focus on learning from human demonstrations, known as imitation learning (IL)\cite{IL2024sun, IL2024wan, IL2024lin, IL2024liu}. This methodology encompasses three primary steps: data collection by experts, the learning model, and execution\cite{IMI2023zhou, IMI2023franzese}. This paper focuses on the data collection phase.

In IL, data acquisition commonly employs teleoperated systems including joysticks, VR controllers, motion capture, and leader-follower systems\cite{tre2018Zhang,tre2021tung,tre2018fan,act2023zhao, mact2024fu, ge2023swu}. Systems such as ALOHA\cite{act2023zhao}, Mobile ALOHA\cite{mact2024fu}, and GELLO\cite{ge2023swu} effectively gather joint angles and images of robots by utilizing a leader and follower setup. However, these systems do not handle force information, thus limiting their ability to discern the hardness of objects. Additionally, data collection in these systems is capped at 50Hz, not meeting the needs for higher sampling rates. This has led to increased interest in methods that handle both positional and force information through bilateral control\cite{IMIB2022sakaino}.

\begin{figure}[t]
  \begin{center}
    \scalebox{0.51}{
        \includegraphics{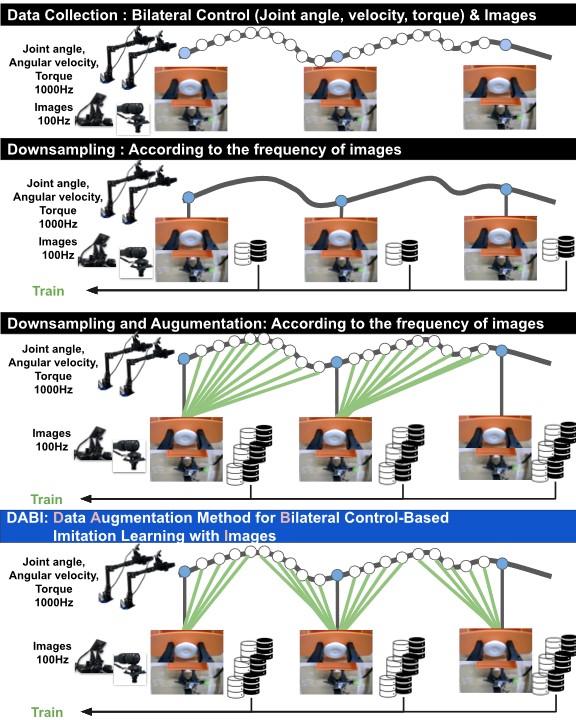}}
        \caption{Overview of Data Augmentation Method for Bilateral Control-Based Imitation Learning with Images (DABI)}
  \label{fig:overview}
\end{center}
\end{figure}

In bilateral control-based imitation learning, the operator controls the leader robot, and the follower robot moves in synchronization with its movements. This control is realized through position tracking and the principle of action-reaction, allowing the operator to directly feel the force and tactile information from the follower robot. This approach has been implemented by Sakaino et al. with data collection at 500Hz and inference periods between 25-50Hz using Long Short-Term Memory (LSTM) networks, which has proven effective in learning and predicting time-series data \cite{lstm, IMIB2022sakaino, IMIB2022hayashi, IMIB2019fujimoto}. Meanwhile, Kobayashi et al. proposed ILBiT utilizing transformer encoder and data collection at 1000Hz to overcome the limitations of LSTM in handling long-term time-series data and generalization\cite{IMIB2023kobayashi}. However, the absence of image information in ILBiT reduced its adaptability to environmental changes.

To address these challenges, Buamanee and Kobayashi et al. developed Bi-ACT by adapting the Action Chunking with Transformers (ACT) approach used in ALOHA and Mobile ALOHA to bilateral control-based imitation learning\cite{IMIB2024Buamanee}. Bi-ACT collects robot joint angles, velocities, and torques at 1000Hz and downsamples the robot data and images to 100Hz to match the model inference period of Bi-ACT. This method has demonstrated its capability to adapt to diverse objects and handle complex tasks, although it requires the collection of 50 demonstrations to learn a single task, similar to the 20-50 demonstrations collected in ALOHA's ACT. Additionally, image data collection with cameras involves different data acquisition intervals (sampling times) for each robot or camera, necessitating preprocessing such as downsampling to match the lowest frequency.

This paper proposes the data augmentation method for bilateral control-based imitation learning with images (DABI), showcasing the approach in Fig.~\ref{fig:overview}. The main contributions are highlighted in two key areas:
\begin{itemize}
\item We collected 5 demonstration datasets using bilateral control and trained the Bi-ACT model with three different variations: with and without data augmentation methods. We conducted real-world experiments on putting items in a drawer, where DABI achieved the highest success rate, demonstrating the effectiveness of the DABI approach. 
\item We discovered that differences in data augmentation methods following downsampling lead to variations in task success rates in bilateral control-based imitation learning.
\end{itemize}
This paper consists of six sections. Sections II and III explain  Related Works and Control System.
Section IV proposes DABI, while Section V confirms the method's efficiency based on the experimental results.
Section VI concludes this paper.

\section{Related Works}
\subsection{Bilateral Control-Based Imitation Learning}
\begin{figure}[t]
  \begin{center}
    \scalebox{0.25}{
        \includegraphics{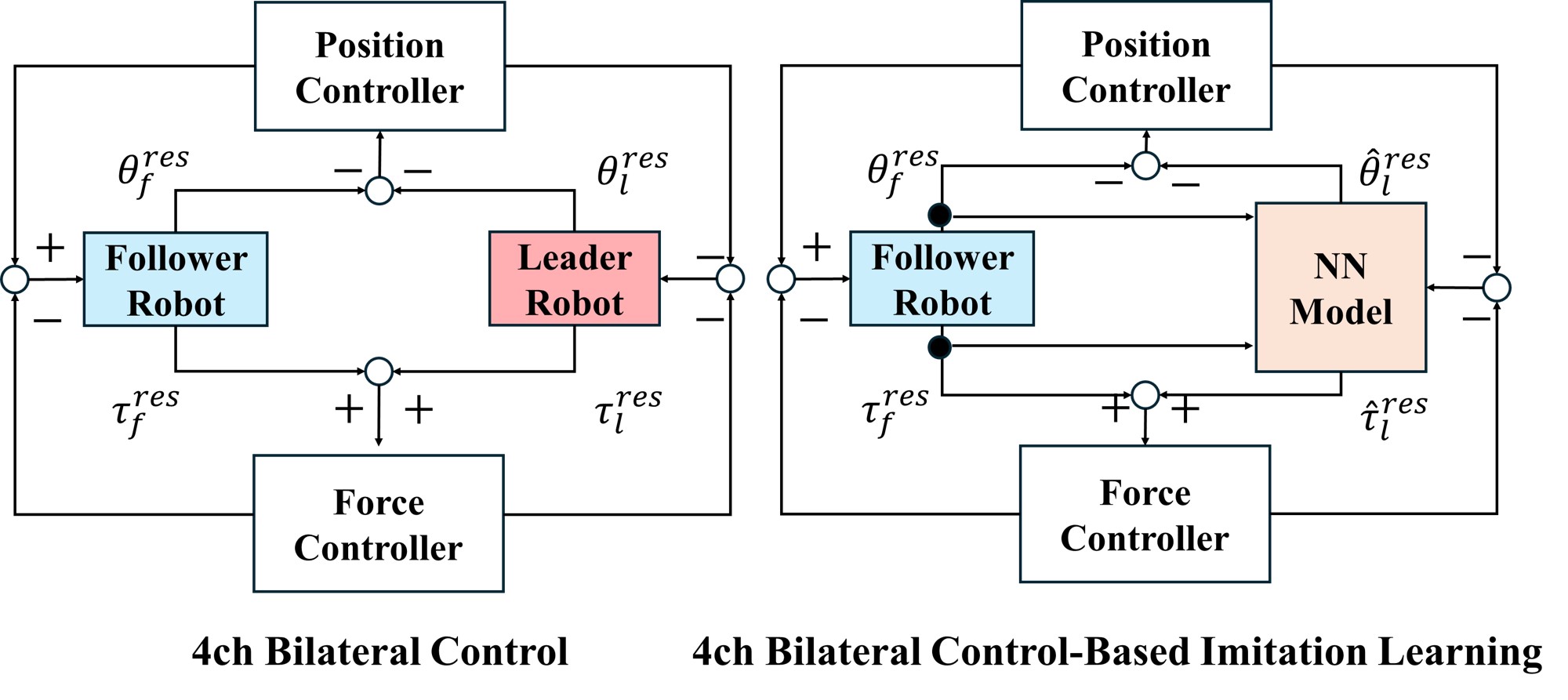}}    
  \caption{Block Diagram of Bilateral Control-Based Imitation Learning}
  \label{fig:4ch}
\end{center}
\end{figure}
As showin Fig.~\ref{fig:4ch}, bilateral control-based imitation learning, capable of collecting position and force information\cite{IMIB2022hayashi}~-\cite{IMIB2023kobayashi}. Bilateral control involves the remote operation of a follower robot in the environment, guided by a leader robot controlled by a human. This is achieved through position tracking and the use of action-reaction principles. Bilateral Control-Based Imitation Learning methods using LSTM have successfully completed various tasks\cite{IMIB2022sakaino, lstm}.
Additionally, Kobayashi et. al. have proposed imitation learning with bilateral control utilizing Transformers\cite{IMIB2023kobayashi,TRANS2017vaswani}.
Buamanee and Kobayashi et al. developed Bi-ACT by adapting the Action Chunking with Transformers (ACT) approach used in ALOHA and Mobile ALOHA to bilateral control-based imitation learning\cite{IMIB2024Buamanee}. Bi-ACT collects robot joint angles, velocities, and torques at 1000Hz and downsamples the robot data and images to 100Hz to match the model inference period of Bi-ACT. This method has demonstrated its capability to adapt to diverse objects and handle complex tasks, although it requires the collection of 50 demonstrations to learn a single task, similar to the 20-50 demonstrations collected in ALOHA's ACT. Additionally, image data collection with cameras involves different data acquisition intervals for each robot or camera, necessitating preprocessing such as downsampling to match the lowest frequency.
Addressing this gap, our study introduces DABI, the data augmentation for bilateral control-based imitation learning with images.

\subsection{Data Augmentation Method for Robot Learning}
Data augmentation methods can generalize robotic skills, enhance success rates, and improve the efficiency of data collection. Yu et. al have proposed "Robot Learning with Semantically Imagined Experience (ROSIE)," which utilizes a text-to-image diffusion model\cite{rosie2023yu}.
By using text guidance, this method performs data augmentation on existing robotic manipulation datasets by inpainting various unseen objects related to manipulations, backgrounds, and obstacles.
In training neural networks with time-series data, learning efficiency can be improved by reducing the sampling frequency to a certain extent \cite{ds2018rah}.
In bilateral control-based imitation learning, our approach closely aligns with that of Yamane et al., who collected robot data at 500 Hz and images at 50 Hz, expanding the data tenfold through downsampling and data augmentation\cite{da2024yamane}.
Methods for data augmentation following downsampling include utilizing data close to timestamps or using the data available from the acquisition of one image until the next image arrives.

A key feature of our DABI approach is focusing on the amount of robot data before and after image acquisition without relying on timestamps, using the data evenly, and performing data augmentation based on the images. This paper also focuses on the evaluation during data augmentation. In this paper, images are collected at 100 Hz and robot data at 1000 Hz, enabling a tenfold data augmentation.

\section{Control System}
\subsection{Controller}
\begin{figure}[t]
  \begin{center}
    \scalebox{0.22}{
        \includegraphics{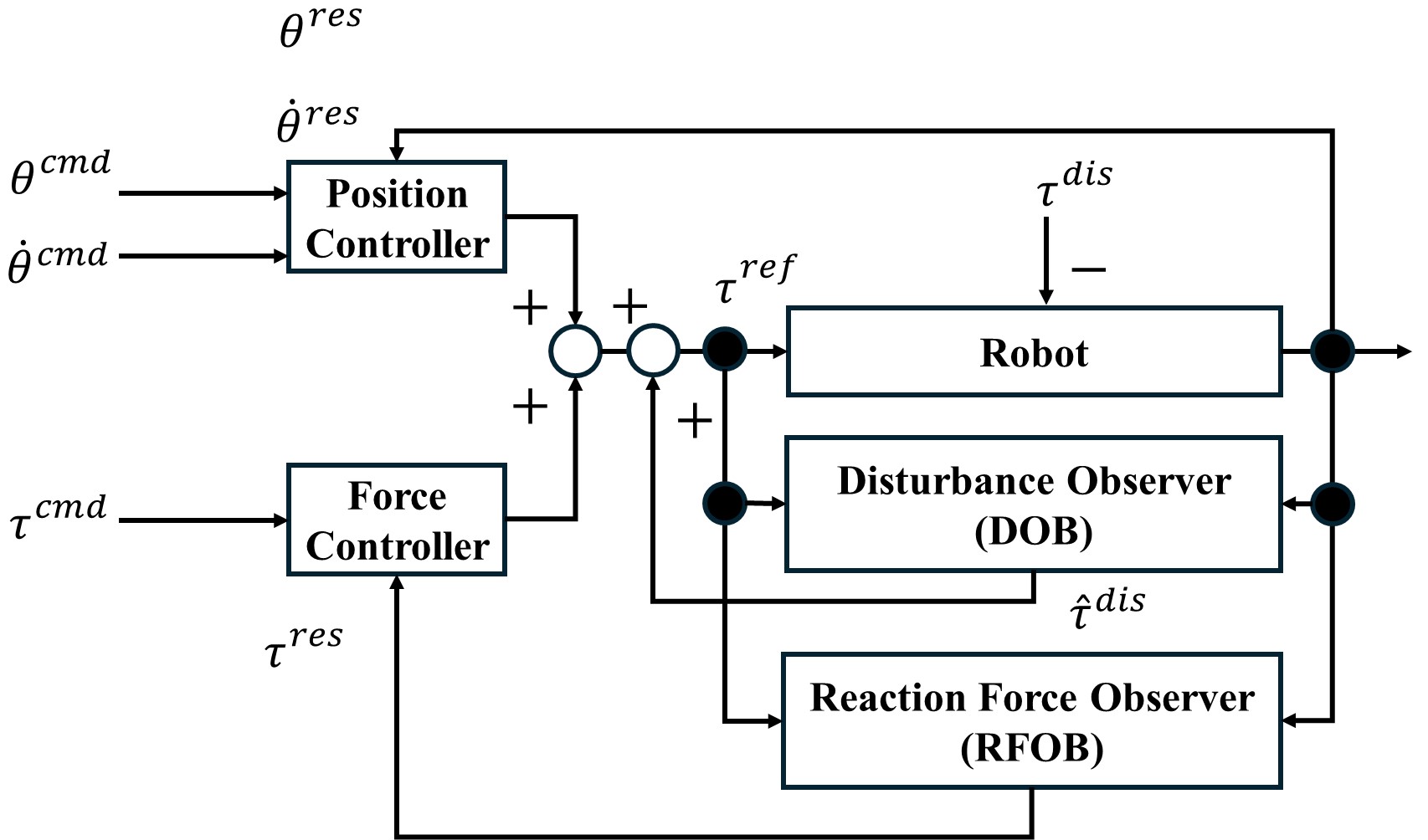}}    
  \caption{Block Diagram of Robot Control System}
  \label{fig:robot_system}
\end{center}
\end{figure}

The controller design adopts a control of position and force for each axis, as shown in Fig.~\ref{fig:robot_system}.
Angle information was obtained from encoders, and angular velocity was calculated by differentiating this information. The disturbance torque $\hat{\tau}^{dis}$ was calculated using a disturbance observer (DOB)\cite{DOB}, and the torque response value $\tau^{res}$ was estimated using a force reaction observer (RFOB)\cite{RFOB}.
\subsection{Bilateral Control}
Bilateral control, characterized by its two-way communication and feedback loop, allows for a more nuanced and synchronized interaction between the leader (human operator) and the follower (robot), essential for delicate manipulation or operation in dynamic environments. The core principle of bilateral control involves sharing position, force, or other relevant information between the operator and the control target. The goals include:
\begin{equation}
\theta_l - \theta_f = 0
\label{eq:position} 
\end{equation}
\begin{equation}
\tau_l + \tau_f = 0
\label{eq:force}
\end{equation}
where $\theta$ and $\tau$ represents the joint angle and torque. The subscript $\bigcirc_l$ represents the leader system, and $\bigcirc_f$ represents the follower system. This allows the operator to perform intuitive control over the control target.
Bilateral control is achieved by satisfying (\ref{eq:position}) for position tracking between systems and (\ref{eq:force}) for the action-reaction relationship of forces.

\section{DABI: Data Augmentation Method for Bilateral Control-Based Imitation Learning with Images}
\begin{figure*}[t]
  \begin{center}
    \scalebox{0.53}{
        \includegraphics{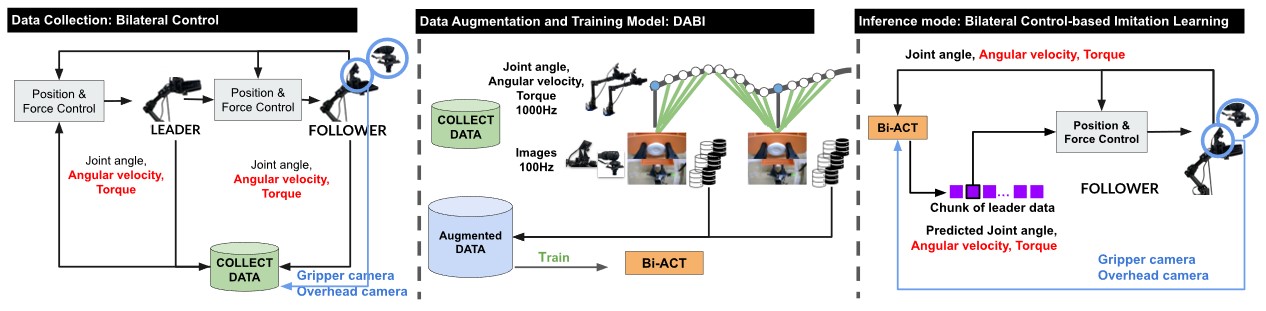}}
  \caption{DABI: Data Augmentation Method for Bilateral Control-Based Imitation Learning with Images}
  \label{fig:OverofDABI}
\end{center}
\end{figure*}
\begin{figure}[t]
  \begin{center}
    \scalebox{0.35}{
        \includegraphics{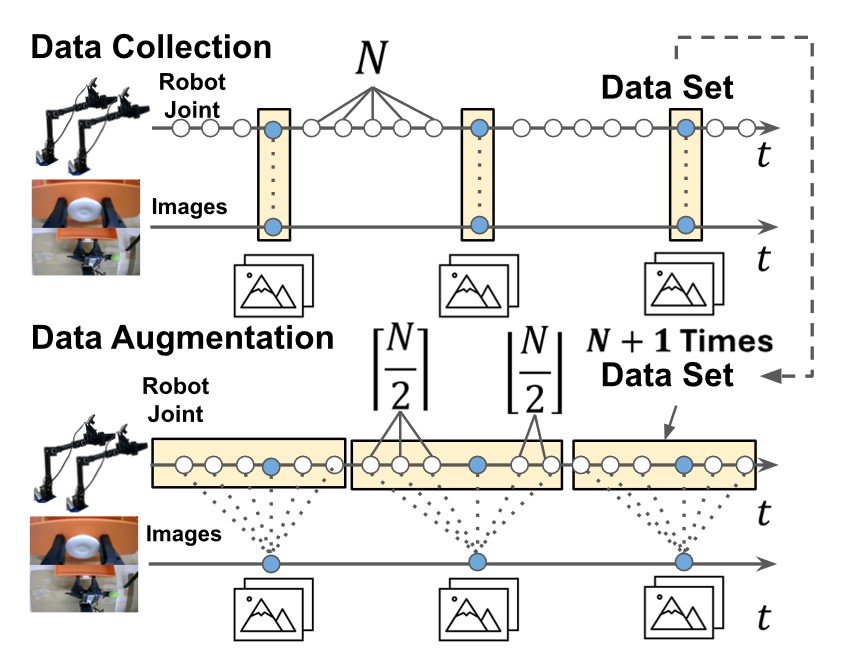}}
  \caption{Image Diagram of Data Augmentation in DABI}
  \label{fig:ImageofDABI}
\end{center}
\end{figure}
\subsection{Overview}
Our proposed research is a data augmentation method for the Bi-ACT learning data. Bi-ACT is inspired by the groundbreaking methodology presented in the ACT study. ACT operates solely on position control using input data of joint angles and images. In contrast, Bi-ACT uses input data of joint angles, velocities, torques, and images to operate with bilateral control that incorporates both position and force control.

Fig.~\ref{fig:OverofDABI} illustrates the overview of the proposed DABI. First, we collect data at 1000 Hz for joint angles, angular velocities, and torques of the leader and follower robots using bilateral control, and images from the gripper and overhead cameras at 100 Hz. Next, the collected data is usually downsampled to match the lower frequency for imitation learning. In this case, it is matched to the 100 Hz of the collected camera images. A feature of DABI is that it uses equidistant robot data before and after the image as a starting point for data augmentation. The augmented data is then used for learning in Bi-ACT.
The trained model of Bi-ACT operates the robot.

\subsection{Data Collection}
Data collection was conducted using Bilateral Control, allowing the user to sense the environment of the follower robot while controlling the leader robot, thereby enhancing the quality of the demonstration data. The joint angles, angular velocities, and torques of both the leader and follower robot arms were collected at 1000 Hz, while images from the overhead and gripper cameras were collected at 100 Hz.

\subsection{Data Augmentation}
Robot operation data and image data are usually recorded at different acquisition cycles, so it is common in imitation learning to downsample other data to match the lower frequency. In this study, we propose a new data augmentation method called DABI (Data Augmentation for Bilateral Control Imitation Learning) and apply it to an imitation learning system based on bilateral control, named "Bi-ACT." This method assumes that the robot data and image data are synchronized and acquired at a fixed interval.

As shown in Fig.~\ref{fig:ImageofDABI}, DABI defines a set of robot data collected between the cycles of image acquisition as $N$. Using this $N$, DABI forms pairs of robot data symmetrically around the image and can expand the data by $N+1$ times. Ceiling and floor functions are used to properly adjust the amount of data in DABI.

This method allows for the creation of a richer learning dataset while maintaining synchronization between data acquired at different cycles.

\section{Experiments}
\subsection{Hardware}
\begin{figure}[t]
  \begin{center}
    \scalebox{0.25}{
\includegraphics{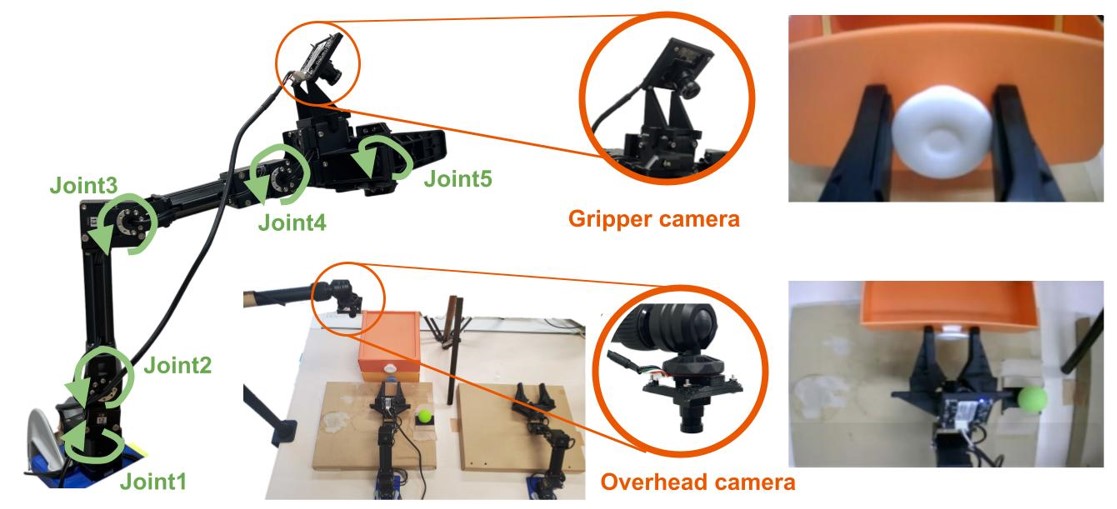}}  
  \caption{Definition of Robot and Camera View}
  \label{fig:Hardware}
\end{center}
\end{figure}
\begin{figure}[t]
  \begin{center}
    \scalebox{0.32}{
        \includegraphics{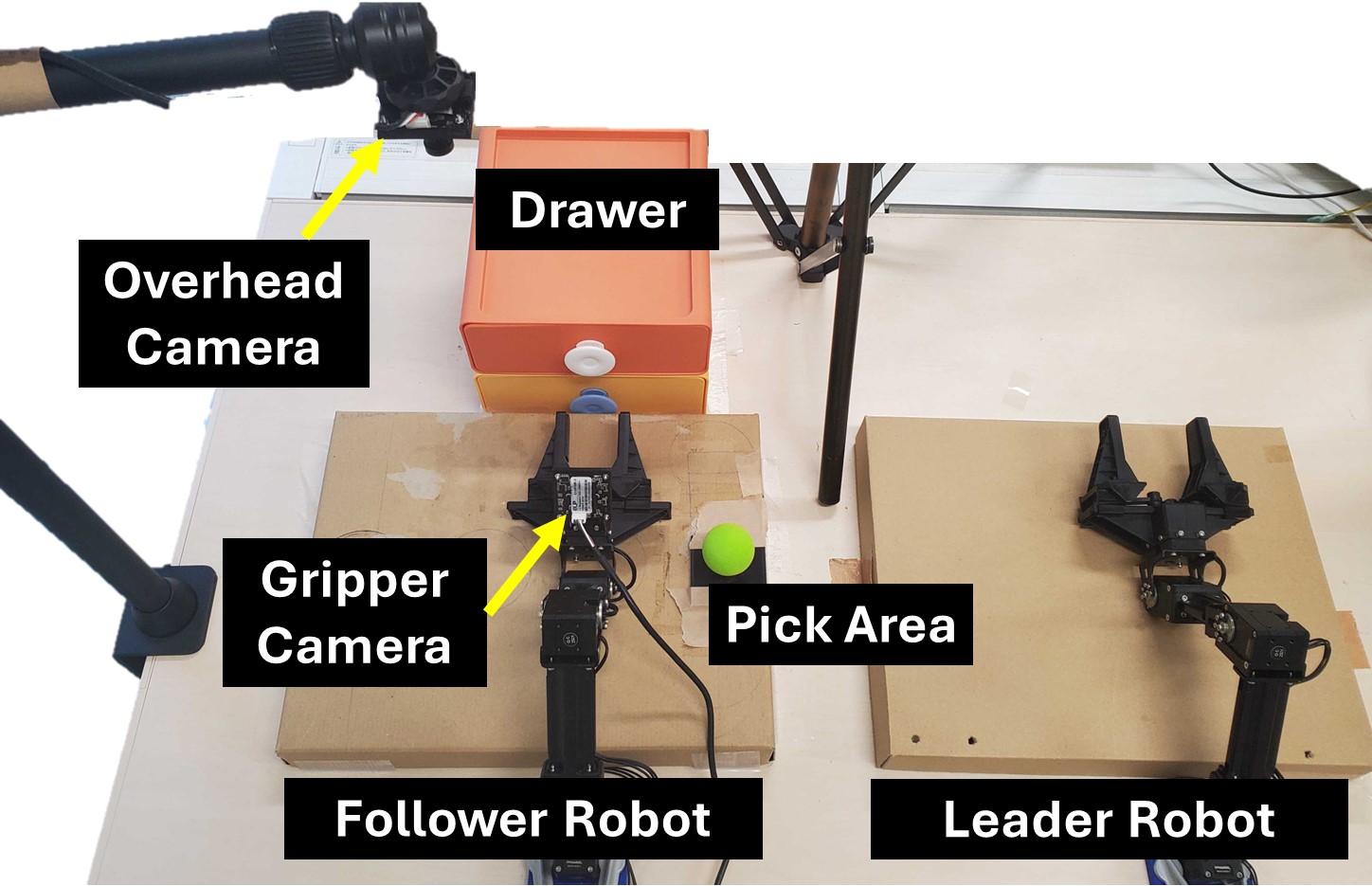}}    
  \caption{Experimental Environments Setup}
  \label{fig:Environment-Setup}
\end{center}
\end{figure}
As shown in Fig.~\ref{fig:Hardware}, experiments were conducted using the OpenMANIPULATOR-X robotic arm by ROBOTIS, featuring 4 Degrees of Freedom (DOF), enabling movement in multiple directions, and an additional DOF for the gripper. The control cycle was set to 1000Hz for precise movement. Moreover, RGB cameras were placed in the overhead and on the gripper area of the follower robot to record observations.

\subsection{Environment Setting}
In the environment setup, two robotic units, the leader and the follower, were positioned adjacent to each other as shown in Fig.~\ref{fig:Environment-Setup}. The experimental environment was arranged on the side of the follower robot. We evaluated DABI's performance on the manipulation task: Put-in-Drawer. The task was performed to test the performance of long-duration tasks.
In the Put-in-Drawer task, the first step is to open the drawer. Next, the robot arm grabs an object and carries it to the drawer. The task then involves placing the object inside the drawer and closing it.

\subsection{Learning Architecture}
\begin{figure}[t]
  \begin{center}
    \scalebox{0.25}{
        \includegraphics{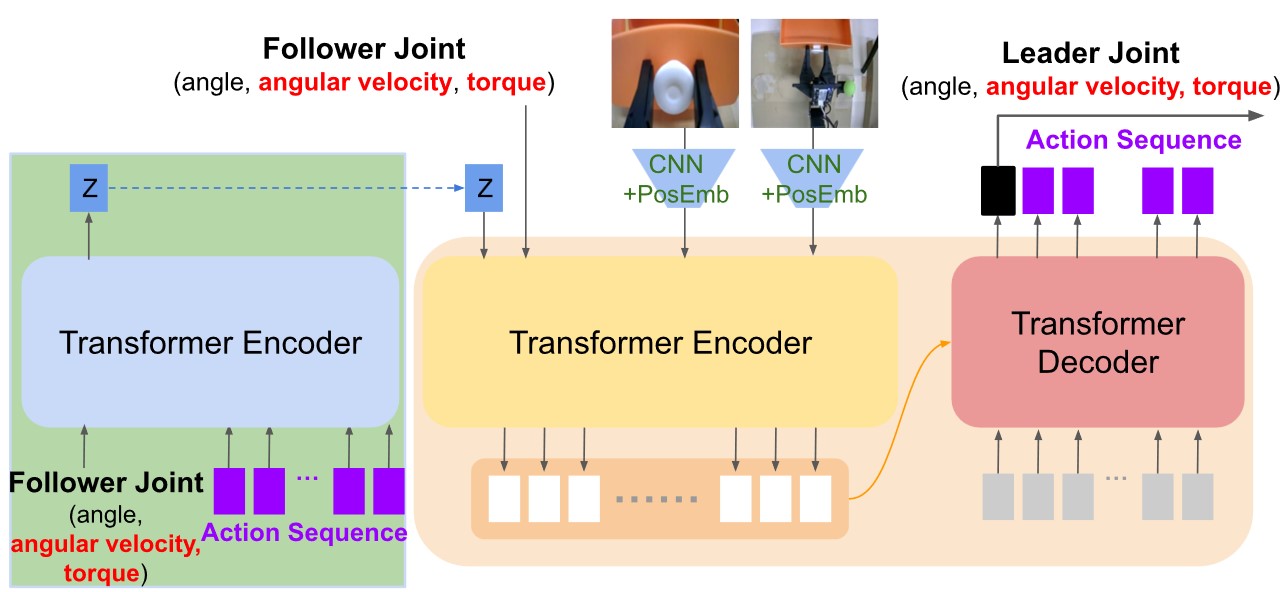}}
  \caption{Model Architecture of Bi-ACT}
  \label{fig:Proposed-Model}
\end{center}
\end{figure}
As shown in Fig.~\ref{fig:Proposed-Model}, the Bi-ACT\cite{IMIB2024Buamanee} model architecture incorporates two RGB images at a 360 x 640 resolution, one from the follower's gripper and the other from an overhead view. In addition, the model processes the current follower's 5 joints, which consist of three types of data (angle, angular velocity, and torque) forming a 15-dimensional vector. Utilizing action chunking, the policy outputs a $k$ x 15 tensor, representing the leader's next actions over $k$ time steps. These actions are then relayed to the controller, which calculates the current needed for the follower's joints to execute the specified movements.

\begin{figure}[t]
  \begin{center}
    \scalebox{0.3}{
        \includegraphics{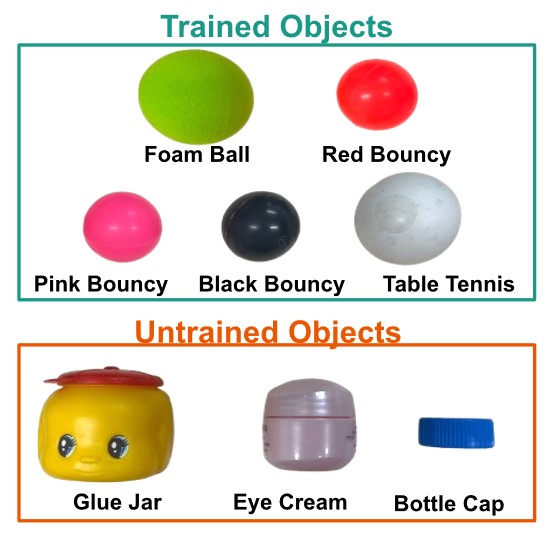}}    
  \caption{Image of Trained and Untrained Objects}
  \label{fig:Target-Object}
\end{center}
\end{figure}

\begin{table}[t]
  \begin{center}
  \caption{Detail of Trained and Untrained Objects}
    \scalebox{1}{
\begin{tabular}{cccc}

\hline
 &   Training Data & Size [mm] &  Weight [g]\\ \hline \hline
Foam ball  & \ding{52} & 40 & 3 \\ 
Red Bouncy  & \ding{52} & 30 & 14 \\ 
Pink Bouncy  & \ding{52} & 30 & 14 \\ 
Black Bouncy  & \ding{52} & 30 & 14 \\ 
Table tennis  & \ding{52} & 40 & 2 \\ 
Bottle Cap   & \ding{54} & 29 & 1 \\ 
Eye cream   & \ding{54} & 40 & 24 \\ 
Glue Jar   & \ding{54}& 45 & 63 \\ \hline
\end{tabular}
}
\label{tab:data_obj}
\end{center}
\end{table}

\begin{table*}[t]
    \centering
    \caption{Experimental Results: Put-in-Drawer}
    \scalebox{1}{
    \begin{tabular}{c|cc|c|c|c|c|c|c}
        \hline
        \multicolumn{1}{c}{\multirow{2}{*}{Method}} & \multicolumn{2}{c}{\multirow{2}{*}{Objects}} &\multicolumn{6}{c}{Put-in-Drawer}\\
        \multicolumn{1}{c}{}&\multicolumn{1}{c}{}&\multicolumn{1}{c}{}&\multicolumn{1}{c}{Open}&\multicolumn{1}{c}{Pick}&\multicolumn{1}{c}{Move}&\multicolumn{1}{c}{Place}&\multicolumn{1}{c}{Close}&Total\\\hline \hline

        \multirow{8}{*}{Method~1} &\multirow{5}{*}{Trained}&Foam ball&60&40&40&40&40&40\\
        \multirow{8}{*}{Downsampling without Augmentation} &&Red bouncy&20&20&20&20&20&20\\
        \multirow{8}{*}{} &&Pink bouncy&0&0&0&0&0&0\\
        &&Black bouncy&0&0&0&0&0&0\\
        &&Table Tennis&40&0&0&0&0&0\\\cline{2-9}
        &\multirow{4}{*}{Untrained}&Bottle cap&40&0&0&0&0&0\\
        &&Eye Cream&0&0&0&0&0&0\\\
        &&Glue Jar&40&40&40&40&40&40\\\hline

        \multirow{8}{*}{Method~2}
        &\multirow{5}{*}{Trained}&Foam ball&100&100&100&100&100&100\\
        \multirow{8}{*}{Downsampling and Augmentation} 
        &&Red bouncy&60&60&60&60&60&60\\
        \multirow{8}{*}{}&&Pink bouncy&80&80&80&80&80&80\\
        &&Black bouncy&100&80&80&80&80&80\\
        &&Table Tennis&80&80&80&80&80&80\\\cline{2-9}
        &\multirow{4}{*}{Untrained}&Bottle cap&40&40&40&40&40&40\\
        &&Eye Cream&80&80&80&80&80&80\\\
        &&Glue Jar&40&40&40&40&40&40\\\hline

        \multirow{8}{*}{Method~3}&\multirow{5}{*}{Trained}&Foam ball&100&100&100&100&100&100\\
         \multirow{8}{*}{DABI} &&Red bouncy&80&80&80&80&80&80\\
        \multirow{8}{*}{}&&Pink bouncy&100&100&100&100&100&100\\
        &&Black bouncy&80&80&80&80&80&80\\
        &&Table Tennis&100&100&100&100&100&100\\\cline{2-9}
        &\multirow{4}{*}{Untrained}&Bottle cap&100&100&100&100&100&100\\
        &&Eye Cream&100&100&100&100&100&100\\\
        &&Glue Jar&100&100&100&100&100&100\\\hline
    \end{tabular}
    }
    \label{tab:ER}
\end{table*}

\begin{figure}[t]
  \begin{center}
    \scalebox{0.4}{
        \includegraphics{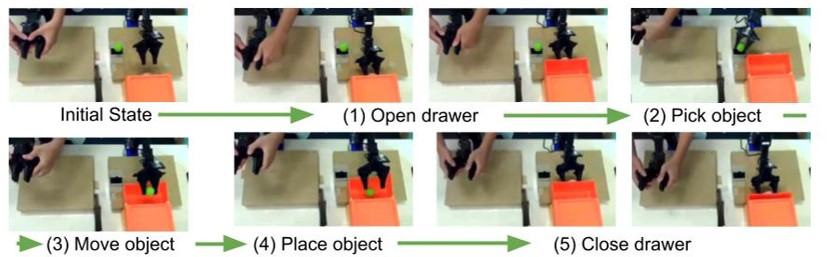}}    
  \caption{Snapshot of Data Collection in Put-in-Drawer}
  \label{fig:data_cllection}
\end{center}
\end{figure}
\subsection{Training Dataset}
Joint angles, angular velocities, and torques for a Leader-Follower robot's demonstration are collected using a bilateral control system. The robot was controlled at a frequency of 1000Hz. Additionally, both gripper and overhead cameras were operating at approximately 100Hz. To synchronize with the system’s operating cycle, the data was adjusted to 100Hz, matching the model’s inference cycle.

For the Put-in-Drawer task, we collected one demonstration for each of the 5 types of training objects as shown in Fig.~\ref{fig:Target-Object} and Table \ref{tab:data_obj}. Using this collected data, we created three types of learning dataset methods as shown in Fig.~\ref{fig:overview}.
The first dataset Method1, involved downsampling the robot data to match the frequency of the camera images at 100Hz and consists only of data from these 5 demonstrations.
The second dataset Method2, expanded the 5 demonstrations from Method1 into 50 demonstrations through data augmentation. This method uses robot joint data starting at the same time as the captured image and continues until the next image is captured.
The third dataset, proposed Method3, also expands the 5 demonstrations from Method1 into 50 demonstrations using data augmentation. This method differs by using the DABI technique, which starts with the image and uses equidistant data before and after the image capture to expand the dataset.

All learning datasets were trained using the Bi-ACT model.
\subsection{Experimental Results}
\begin{figure}[t]
  \begin{center}
    \scalebox{0.48}{
        \includegraphics{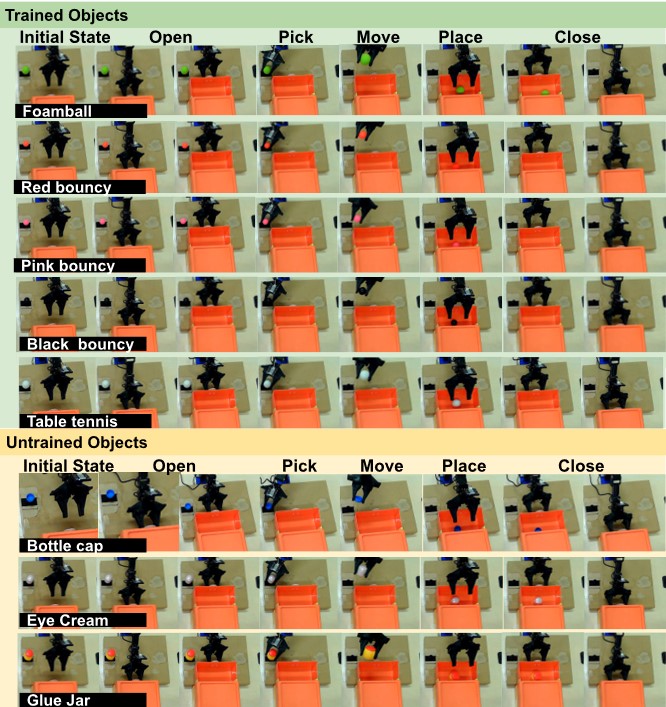}}    
  \caption{Experiment Result: Put-in-Drawer (Proposed Method)}
  \label{fig:case2}
\end{center}
\end{figure}
The results of our experimental evaluations performed on the Put-in-Drawer task using three different methodologies: downsampling(Method1), downsampling with augmentation(Method2), and DABI(Method3). The task performance was assessed across various stages: Open, Pick, Move, Place, and Close, with results compiled in Table \ref{tab:ER}. Each object was evaluated five times.

As shown in Table\ref{tab:ER}, the Method1 showed limited effectiveness, particularly with untrained objects where most failed to proceed beyond the opening stage. Notably, trained objects like the Foam Ball showed mediocre performance, while others like the Pink and Black bouncy balls failed to perform any task successfully.

The introduction of data augmentation in the Method2 marked a significant improvement, especially for trained objects such as the Foam ball, which achieved a perfect success rate across all tasks. However, while performance increased for untrained objects like the Eye Cream, they still showed inconsistency in handling different stages of the task.

The DABI method demonstrated superior performance with a consistent 100\% success rate across all objects and stages. This method proved highly effective not only for trained but also for untrained objects, showcasing the robot's improved adaptability and robust handling capabilities.

The results confirmed a significant improvement in success rates, thereby proving the effectiveness of DABI.

\section{Conclusion}
This paper proposed the data augmentation method for imitation learning called "DABI: Data Augmentation Method for Bilateral Control-Based Imitation Learning with Images".
DABI collected robot joint angles, velocities, and torques at 1000 Hz, and used images from gripper and environmental cameras captured at 100 Hz as the basis for data augmentation.
This allowed for a tenfold increase in data augmentation compared to downsampling methods. Using DABI, we collected 5 demonstration datasets, trained using the Bi-ACT model with bilateral control, and conducted real-world experiments.
The results confirmed a significant improvement in success rates. The effectiveness of this approach was confirmed through real-world experiments.

The limitations and future challenges of this paper are as follows. Even though the experiments used different objects, they were only tested on the Put-in-Drawer task. In the future, we will run experiments on a wider range of tasks. For data augmentation, we used floor and ceiling functions. However, we plan to compare this method with one that matches the closest timestamps between image and robot data.


\end{document}